\documentclass[sigconf]{acmart}
\usepackage{algorithm}
\usepackage{algorithmic}
\usepackage{enumitem}
\usepackage{subcaption}
\usepackage{multirow}
\usepackage{wrapfig}
\usepackage{dblfloatfix}
\usepackage{tablefootnote}
\usepackage{fixltx2e}
\PassOptionsToPackage{bookmarks=false}{hyperref}

\usepackage{tikz}

\newcommand*\circled[1]{\tikz[baseline=(char.base)]{
            \node[shape=circle,draw,inner sep=0.8pt, minimum size=2pt] (char) {#1};}}
            
\newcommand{\rpoint}[1]{\circled{{\fontfamily{pcr}\selectfont\footnotesize{#1}}}}

\AtBeginDocument{%
  \providecommand\BibTeX{{%
    \normalfont B\kern-0.5em{\scshape i\kern-0.25em b}\kern-0.8em\TeX}}}

\acmBooktitle{Woodstock '18: ACM Symposium on Neural Gaze Detection,
  June 03--05, 2018, Woodstock, NY}
\acmPrice{15.00}
\acmISBN{978-1-4503-XXXX-X/18/06}

\usepackage{fancyhdr,lipsum}
\fancypagestyle{firstpage}{% Page style for first page
  \fancyhf{}% Clear header/footer
  \fancyhead[C]{To appear at the IEEE/ACM International Conference on Computer-Aided Design (ICCAD '20), November 2--5, 2020, Virtual Event, USA.}% Header
  \fancyfoot[C]{\thepage}% Footer
}

\renewcommand\footnotetextcopyrightpermission[1]{} % removes footnote with conference information in first column
\begin{document}

\sloppy
\title{NASCaps: A Framework for Neural Architecture Search\\
to Optimize the Accuracy and Hardware Efficiency of Convolutional Capsule Networks\\
% \begin{normalfont}
% \begin{large}
% Alberto Marchisio$^{*1}$\thanks{$^*$These authors contributed equally}, Andrea Massa$^{*2}$, Vojtech Mrazek$^3$, Beatrice Bussolino$^{2}$, Maurizio Martina$^2$, Muhammad Shafique$^{1,4}$
% \end{large}\\
% \vspace*{-0pt}
% \begin{small}
% \{alberto.marchisio, muhammad.shafique\}@tuwien.ac.at, a.massa@studenti.polito.it, mrazek@fit.vutbr.cz, \{beatrice.bussolino, maurizio.martina\}@polito.it
% \\
% \vspace*{-0pt}
% \textit{\textsuperscript{1}Institute of Computer Engineering, Technische Universit\"{a}t Wien, Austria, \hspace{0.3cm} \textsuperscript{2}Department of Electronics and Telecommunications, Politecnico di Torino, Italy,}\\
% \vspace*{-10pt}
% \textit{\textsuperscript{3}Faculty of Information Technology, Brno University of Technology, Czech Republic, \hspace{0.3cm} \textsuperscript{4}Division of Engineering, New York University Abu Dhabi, UAE}
% \end{small}
% \end{normalfont}
\vspace*{10pt}
}

\author{{Alberto Marchisio$^{*1}$, Andrea Massa$^{*2}$, Vojtech Mrazek$^3$, Beatrice Bussolino$^{2}$, Maurizio Martina$^2$, Muhammad Shafique$^{1,4}$}}

%\authornote{These authors contributed equally.}
\affiliation{\vspace*{10pt}\textit{\textsuperscript{1}Institute of Computer Engineering, Technische Universit\"{a}t Wien, Vienna, Austria}}
\affiliation{\textit{\textsuperscript{2}Department of Electronics and Telecommunications, Politecnico di Torino, Turin, Italy}}
\affiliation{\textit{\textsuperscript{3}Faculty of Information Technology, Brno University of Technology, Brno, Czech Republic}}
\affiliation{\textit{\textsuperscript{4}Division of Engineering, New York University Abu Dhabi, UAE}}
\affiliation{\vspace*{10pt}alberto.marchisio@tuwien.ac.at, a.massa@studenti.polito.it, mrazek@fit.vutbr.cz,\\
\{beatrice.bussolino, maurizio.martina\}@polito.it, muhammad.shafique@nyu.edu\vspace*{20pt}}

\thanks{$^*$These authors contributed equally}
\begin{abstract}
%@alberto
\begin{small}

Deep Neural Networks (DNNs) have made significant improvements to reach the desired accuracy to be employed in a wide variety of Machine Learning (ML) applications. Recently the Google Brain's team demonstrated the ability of \textit{Capsule Networks (CapsNets)} to encode and learn spatial correlations between different input features, thereby obtaining superior learning capabilities compared to traditional (i.e., non-capsule based) DNNs. However, designing CapsNets using conventional methods is a tedious job and incurs significant training effort. Recent studies have shown that powerful methods to automatically select the best/optimal DNN model configuration for a given set of applications and a training dataset are based on the \textit{Neural Architecture Search (NAS)} algorithms. Moreover, due to their extreme computational and memory requirements, DNNs are employed using the specialized hardware accelerators in IoT-Edge/CPS devices.  

In this paper, we propose \textbf{NASCaps}, an automated framework for the hardware-aware NAS of different types of DNNs, covering both traditional convolutional DNNs and CapsNets. We study the efficacy of deploying a multi-objective Genetic Algorithm (e.g., based on the NSGA-II algorithm). The proposed framework can jointly optimize the network accuracy and the corresponding hardware efficiency, expressed in terms of energy, memory, and latency of a given hardware accelerator executing the DNN inference. Besides supporting the traditional DNN layers (such as, convolutional and fully-connected), our framework is the first to model and supports the specialized capsule layers and dynamic routing in the NAS-flow. We evaluate our framework on different datasets, generating different network configurations, and demonstrate the tradeoffs between the different output metrics. We will open-source the complete framework and configurations of the Pareto-optimal architectures at \url{https://github.com/ehw-fit/nascaps}.
\vspace*{0pt}
\end{small}
\end{abstract}

\keywords{Deep Neural Networks, DNNs, Capsule Networks, Evolutionary Algorithms, Genetic Algorithms, Neural Architecture Search, Hardware Accelerators, Accuracy, Energy Efficiency, Memory, Latency, Design Space, Multi-Objective, Optimization.
}


\maketitle
\thispagestyle{firstpage}

\pagestyle{empty}

\vspace*{-2pt}

\section{Introduction}
%@alberto

%\red{applications for CNNs, introduction of CapsNets, need of NAS to get more automation in the DNN architecture design}

\begin{figure}[h]
    \centering
    \vspace*{4pt}
    \includegraphics[width=0.8\linewidth]{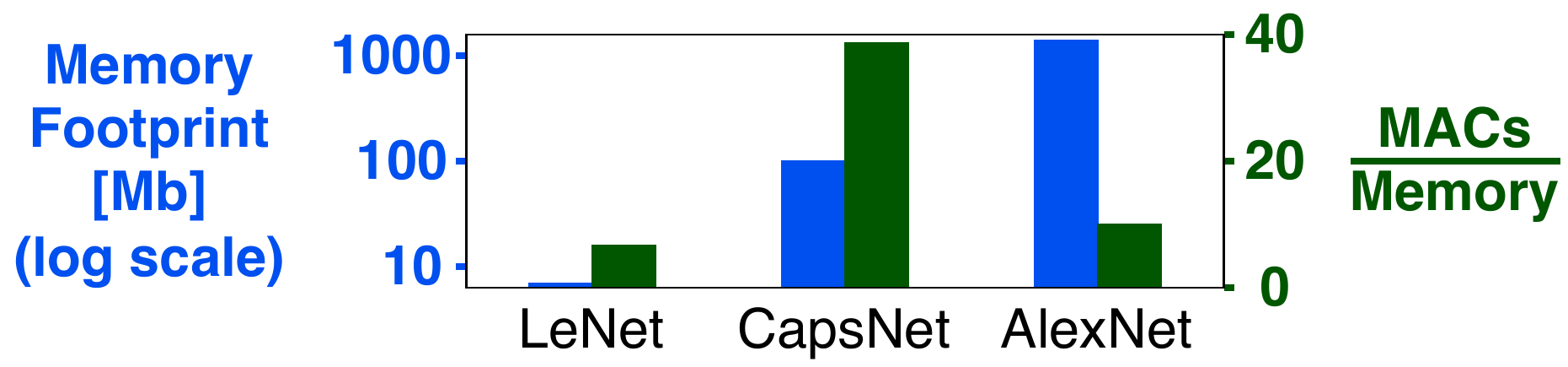}
    \caption{Memory footprint and (Multiply-and-Accumulate operations vs. memory) ratio (MACs/Memory) for the LeNet~\cite{MNIST}, CapsNet~\cite{sabour} and AlexNet~\cite{AlexNet}.}
    \label{fig:mac_memory}
    \vspace*{0pt}
\end{figure}

Deep Neural Networks (DNNs) are advanced machine learning-based algorithms that claim to surpass the human-level accuracy in a certain set of tasks, such as image classification, object recognition, detection, and tracking, when extensively trained over large datasets~\cite{Grigorescu2019surveyAD}\cite{He2015SurpassingHuman-Level}\cite{Redmon2016YOLO}. 
Designing the best DNN for a given set of applications and a given dataset is an extremely challenging effort. Highly-accurate DNNs require a significant amount of hardware resources, which may not be feasible on resource-constrained IoT-Edge devices~\cite{Capra2020Updated}\cite{Marchisio2019DL4EC}\cite{Shafique2018NextGenML}\cite{Sze2017EfficientDNN}. The advanced DNN models, called Capsule Networks (CapsNets)~\cite{sabour}, are able to learn the hierarchical and spatial information of the input features in a closer manner to our current understanding of the human brain's functionality. However, the layers made of capsules introduce an additional dimension w.r.t. the matrices of the convolutional and fully-connected layers of traditional Convolutional Neural Networks (CNNs), which, besides the dynamic routing computations, put an extremely heavy computation and communication workload on the underlying hardware~\cite{capsacc}. In Fig.~\ref{fig:mac_memory} we compare the CapsNet~\cite{sabour} with the LeNet~\cite{MNIST} and the AlexNet~\cite{AlexNet}, and analyze their memory footprints and total number of multiply-and-accumulate (MAC) operations required to perform an inference pass. Note, the MACs/memory ratio is a good indicator to show the computational complexity of the network, thus demonstrating the higher compute-intensive nature of CapsNets, compared to traditional CNNs.

%\vm{from the project proposal ; maybe can be moved to Sec 1.1}
%In the literature, many different architectures have already been published \red{REF-some archits}. In the pioneering age of deep learning, the architectures has been designed manually. However, the structure of neural networks becomes to be very complex. Therefore, automated neural architecture search (NAS) methodologies were proposed. A lot of these methodologies are based on so-called \textit{evolutionary algorithms}  \red{https://arxiv.org/pdf/1808.03818.pdf}. The evolutionary algorithms (EAs) were used for the learning of small NNs. However, in the area of complex DNNs, the EA has already been used for improvement of some parameters of single layers~\cite{V. Mrazek, Z. Vasicek, L. Sekanina, M. A. Hanif: ALWANN:} as well as global hyperparameters~\cite{abs/1703.00548}. Another approach was to optimize the structure (connection) of submodules~\cite{iccv  L. Xie,  and  A. Yuille. (2017). “Genetic CNN”}. But the researchers typically focused on the performance at standard computing platforms such as number of trainable parameters and FLOPS parameters on GPUs. 

\subsection{Research Problem \& Associated Challenges}

%\red{We want to design a NAS methodology including also the possibility to have Capsule Layers, and also include the hardware efficiency metrics as optimization goal. As per our knowledge, we are the first... Challenges are that we need to model the capsule layers in a parametric way and we need to model the hardware parameters of the CapsNet accelerator. Afterwards, we say that other big challenges are the explosion of the search space and huge training time, that can be reduce by estimating (when? at which epoch?) and using a genetic algorithm.}
In the literature, many DNN models/architectures have been proposed~\cite{He2015ResNet}\cite{AlexNet}\cite{Simonyan2014VGG}\cite{inceptionv4}\cite{Szegedy2015GoogleNet}. In the pioneering age of deep learning, the DNN architectures were designed manually. However, their structures became very complex. Therefore, Neural Architecture Search (NAS) methodologies emerged as an attractive procedure to select the optimal DNN model for a given set of applications and a training dataset~\cite{Zoph2017NAS}. Evolutionary Algorithm (EA) based methodologies~\cite{Sun2020GeneticCNN} were proposed for learning small DNNs. However, for more complex DNNs, EAs have been used for improving certain parameters of single/individual DNN layers~\cite{Mrazek2019ALWANN}, as well as global DNN hyperparameters~\cite{Miikkulainen2017EvolvingDNN}, or the connection between submodules~\cite{Xie2017GeneticCNN}.

Most of the automatic tools based on a NAS algorithm that have been proposed in the literature only focus on optimizing the DNN accuracy~\cite{Sun2020GeneticCNN}\cite{Zoph2018nasnet}. Only a few of them have recently introduced the hardware constraints in the optimization problem, for instance, considering the hardware resources (e.g., \#FLOPs, memory requirements, etc.) available for performing the DNN inference~\cite{Jiang2019FPGA-NAS}\cite{Jiang2020co-NAS}\cite{Lu2020MUXConv}\cite{Stamoulis2019SinglePathNAS}. To the best of our knowledge, none of them include in the design space the possibility of employing capsule layers and dynamic routing, which are inevitable for automatically designing the CapsNets.

Toward this, we propose \textbf{NASCaps}, a framework for the NAS of DNNs, that not only incorporates the most common types of DNN layers (such as convolutional, fully-connected) but also, \textit{for the first time, the different types of capsule layers}. Our framework supports multi-objective hardware-aware optimizations because it investigates the network accuracy, and it accounts for different hardware efficiency parameters (such as memory usage, energy consumption, and latency) that are crucial for embedded DNN inference accelerators. 

However, the huge variety of possible configurations that should be explored to obtain an exhaustive set of Pareto-optimal solutions might dramatically explode. In addition to this, despite adopting the most advanced learning policies and employing high-end GPU clusters, complex CapsNets and CNNs typically require long training time~\cite{Harlap2018PipeDream}\cite{Marchisio2019XTrainCaps}. Complete detailed post-synthesis hardware measurements are not feasible for this search due to their long simulation times. The above-discussed limitations challenge the applicability of such an exploration in real-case HW/SW co-design searches, with stringent time-to-market constraints. 

\vspace*{-8pt}
\subsection{Our Key Research Contributions}

\begin{figure}[h]
    \centering
    \includegraphics[width=\linewidth]{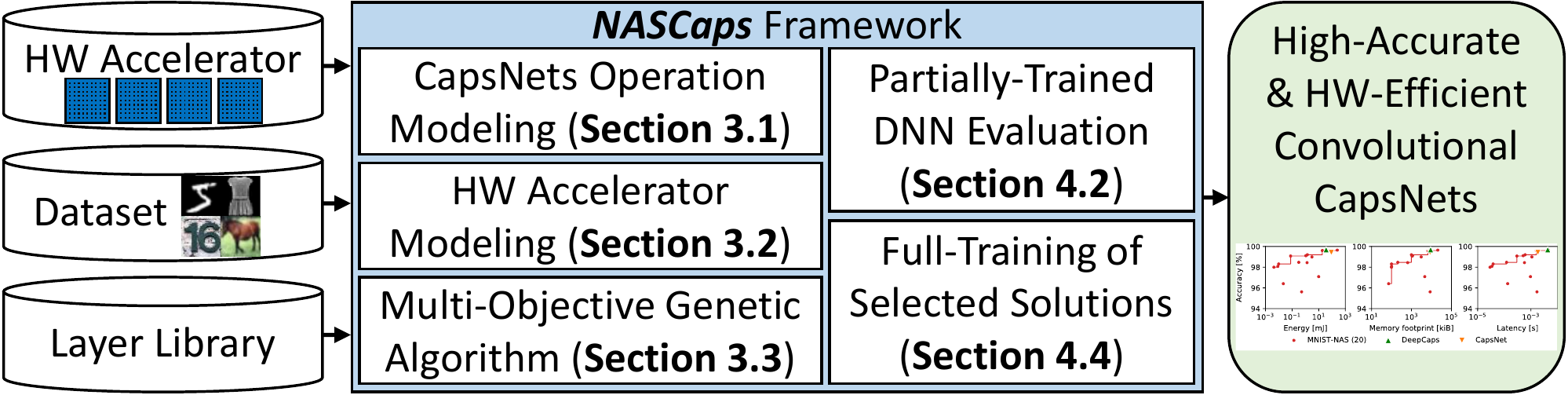}
    \caption{Overview of our \textit{NASCaps} framework.}
    \label{fig:novel_contrib}
    \vspace*{-2pt}
\end{figure}

To address the above challenges, we devise different optimizations and integrate them into our \textit{NASCaps} framework (Fig.~\ref{fig:novel_contrib}). The steps are summarized in the following \textbf{novel contributions}:
\vspace*{-3pt}
\begin{itemize}[leftmargin=*]
    \item We present a framework, called \textit{NASCaps}, to automatically search the DNN model architecture configurations, based on convolutional layers and capsule layers. (\textbf{Section~\ref{sec:main_technical}})
    \item We model the operations involved in the CapsNet architectures in a parametric way, including the different types of capsule layers and the dynamic routing. (\textbf{Section~\ref{subsec:CapsNet_model_SW}})
    \item We model the functional behavior of a given specialized CNN and CapsNet hardware accelerator at a high level, to quickly estimate the memory usage, energy consumption, and latency, when different DNN architectural models are executed. (\textbf{Section~\ref{subsec:CapsNet_model_HW}})
    \item Based on the NSGA-II method~\cite{deb2002fast}, we developed a specialized multi-objective genetic algorithm for solving the optimization problem targeted in this paper, i.e., a multi-objective Pareto-frontier selection of DNN architectures while optimizing the neural network's accuracy, energy consumption, memory usage, and latency. (\textbf{Section~\ref{subsec:genetic algorithm}})
    \item To reduce the training time for the exploration of different solutions, we propose a methodology to evaluate the accuracy of partially-trained DNNs. The number of training epochs is chosen based on the tradeoff between training time and Pearson correlation coefficient w.r.t. fully-trained DNNs. (\textbf{Section~\ref{subsec:reduce_epochs}})
    \item During the exploration phase, we trained and evaluated more than 600 candidate DNN solutions running on the GPU-HPC computing nodes equipped with four high-end Nvidia V100-SMX2 GPUs. The Pareto-optimal solutions generated by our \textit{NASCaps} framework are competitive w.r.t. the previous SoA accuracy values for CapsNets, i.e., the DeepCaps~\cite{deepcaps}, while improving the corresponding hardware efficiency, thereby opening new avenues towards the deployment of high-accurate DNNs at the edge. (\textbf{Section~\ref{subsec:NASCaps_results}})
    \item Towards encouraging fast advancements in the DNN research community, we will open-source the complete \textit{NASCaps} framework and configurations of the Pareto-optimal architectures at \url{https://github.com/ehw-fit/nascaps}. %\vm{the code is not prepared for that; however we can publish the final NNs (in a Python-code definition)}
\end{itemize}

Before proceeding to the technical sections, we provide a brief overview of the CapsNets and the hardware accelerators executing CapsNet inference. (\textbf{Section~\ref{sec:background}})

%In summary, this work presents a framework, called \textit{NASCaps}, that automatically searches the DNN model architecture configurations (including a combination of convolutional layers and capsule layers as well), with the goal of jointly optimizing the DNN accuracy and its hardware efficiency, when executed in specialized inference accelerators. The output of our framework is a set of Pareto-optimal DNN models, for a given dataset and a given hardware accelerator. Our results demonstrate that a selected subset of solutions not only surpass the accuracy of the DeepCaps, but also improves its hardware efficiency, thereby opening new avenues towards the deployment of high-accurate DNNs at the edge. 

\vspace*{-5pt}

\section{Background: Capsule Networks}
\label{sec:background}
The idea of grouping the neurons to form a \textit{capsule} was first proposed by G. Hinton et al. in~\cite{hinton2011}. The purpose of the capsules is to retain the instantiation probability of an entity or a feature, together with information on its instantiation parameters, such as the position, rotation, or width. On the contrary, traditional neurons can only detect features without any knowledge of their spatial characteristics and inter-correlation.  

The first DNN with capsules, i.e., a \textit{Capsule Network} (CapsNet), was proposed in~\cite{sabour} by the Google Brain's team. In this model, the neurons inside a capsule are arranged into the shape of a vector. Each neuron encodes a spatial parameter of the entity associated with the capsule, and the length of the vector represents the probability that the object is present. The CapsNet in~\cite{sabour} consists of three layers, as shown in Fig.~\ref{fig:capsnet}.a: 

\begin{figure}[b]
    \centering
    \includegraphics[width=0.85\linewidth]{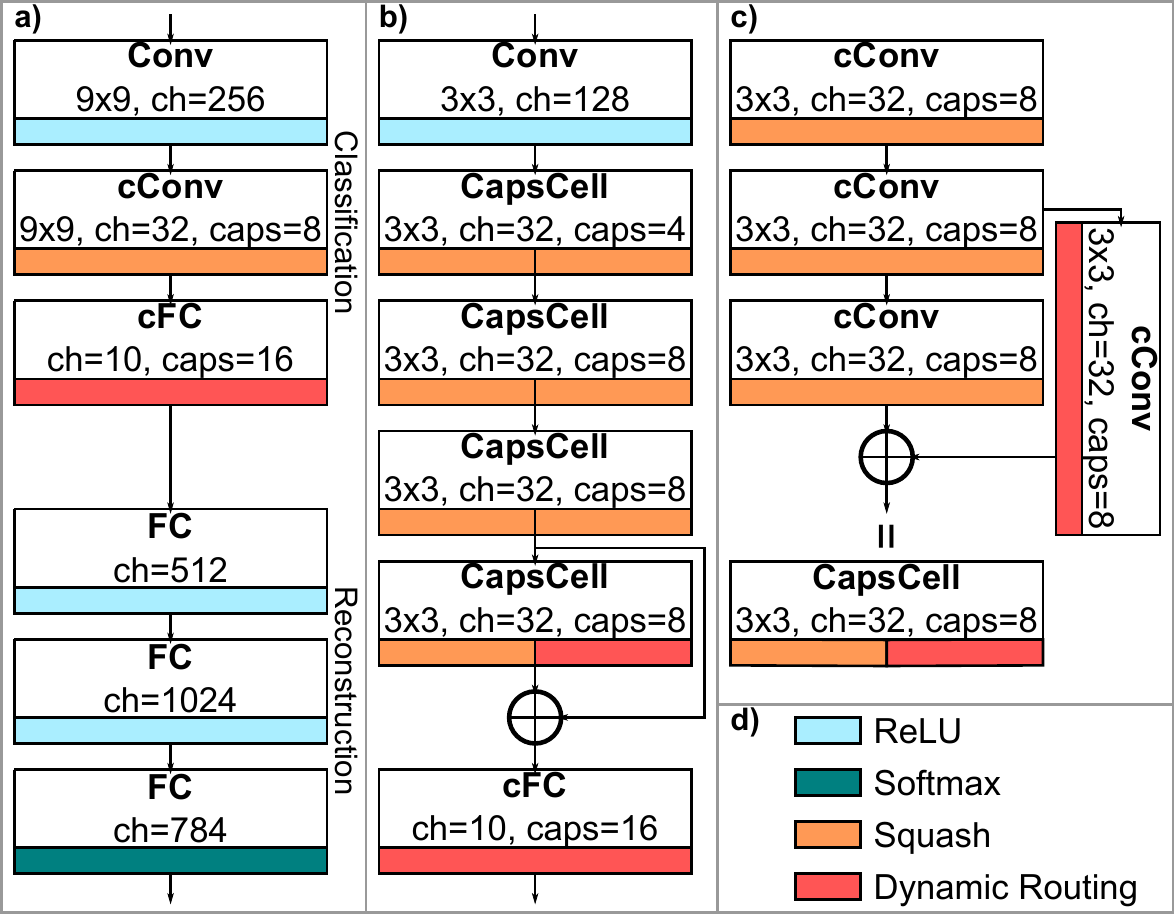}
    \caption{(a) CapsNet model~\cite{sabour} with the decoder for image reconstruction; (b) DeepCaps model~\cite{deepcaps} (the decoder is omitted); (c)~CapsCell used in the DeepCaps model; (d) legend of the activation functions used in each layer. }
    \label{fig:capsnet}
\end{figure}

\begin{enumerate}[leftmargin=*]
\item \textbf{Conv layer}: a convolutional (Conv) layer that applies a 9x9 kernel with stride 1 to the input image and produces 256 output channels. 
\item \textbf{PrimaryCaps layer}: a Conv layer that applies a 9x9 kernel with stride 2 and produces 256 output channels. The output channels are divided into 32 channels of 8-D capsules. Since the length of the capsules encodes a probability, the \textit{squash function} (Eq.~\ref{eq:squash})  is applied to force the length in the range [0,1]. 

\vspace*{-10pt}
\begin{equation}
    \label{eq:squash}
    \mathbf{y} = \frac{|\mathbf{x}|^2}{(1+|\mathbf{x}|)^2}\frac{\mathbf{x}}{|\mathbf{x}|}
\end{equation}

\item \textbf{DigitCaps layer}: a fully-connected (FC) layer of capsules with 10 output 16-D capsules (for a 10-classes dataset). The DigitCaps layer performs the \textit{dynamic routing}, an iterative algorithm that associates coupling coefficients to the predictions obtained from the PrimaryCaps layer. The iterations of the algorithm maximize the coupling coefficients of the capsules predicting the same result (similar spatial parameters) with greater confidence (higher probability of instantiation). 
\end{enumerate}

CapsNets belong to a particular category of DNNs, i.e., \textit{autoencoders}, that can reconstruct the image they receive as input. Therefore, the three layers listed above are followed by a decoder, consisting of three FC layers that reconstruct the input image from the output of the DigitCaps layer. During training, two losses are computed, the loss of the classification network and the reconstruction network. When the inference is performed for classification purposes, the reconstruction network can be removed. To further reduce the processing time and energy, quantization~\cite{Marchisio2020QCapsNets} and approximations~\cite{Marchisio2020ReDCaNe} can be applied to the CapsNets' inference, without decreasing the accuracy much. 

The DeepCaps, a novel deeper CapsNet architecture, has been recently proposed in~\cite{deepcaps}. The architecture, shown in Fig.~\ref{fig:capsnet}.b, is formed by a traditional Conv layer, FC capsule layers (cFC), and Conv capsule layers (cConv). The latter are arranged in blocks, here referred to as \textit{CapsCells} (Fig.~\ref{fig:capsnet}.c), where a cConv layer runs in parallel to two cConvs layers. The decoder used as a reconstruction network consists of a series of de-convolutional layers. The DeepCaps also introduce a skip connection between the last CapsCells to mitigate the vanishing gradient effects that affect deeper networks.

% \begin{figure*}[ht]
% \begin{minipage}{0.6\textwidth}
% \centering
% \includegraphics[width=\linewidth]{img/background_capsnet.pdf}
% \caption{a) CapsNet model~\cite{sabour} with the decoder for image reconstruction; b) legend of the activation functions used in each layer; c) DeepCaps model~\cite{deepcaps} (the decoder is omitted); d) CapsCell used in the DeepCaps model. }
% \label{fig:capsnet}
% \end{minipage}
% \hfill 
% \begin{minipage}{0.26\textwidth}
% \centering
% \includegraphics[width=\linewidth]{img/background_capsacc.pdf}
% \caption{Architectural view of the CapsAcc~\cite{capsacc} accelerator.}
% \label{fig:capsacc}
% \end{minipage}
% \end{figure*}

 \setlength{\columnsep}{10pt}%
 \begin{wrapfigure}{r}{0.6\linewidth}
     \vspace*{-12pt}
   \begin{center}
     \includegraphics[width=\linewidth]{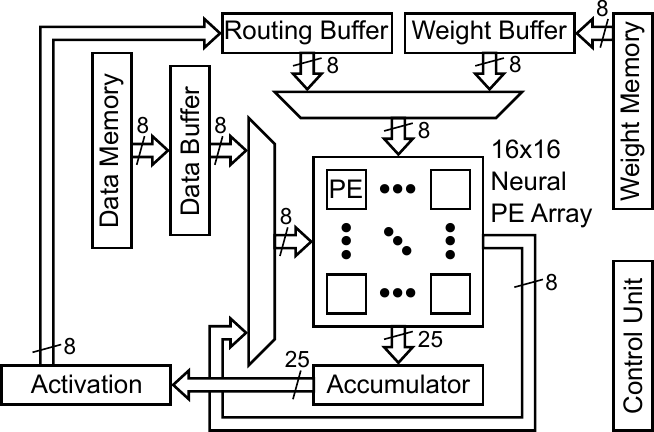}
   \end{center}
   \caption{Architectural view of the CapsAcc~\cite{capsacc} accelerator.}
   \label{fig:capsacc}
   \vspace*{-6pt}
 \end{wrapfigure}

The capsule layers involve operations that are not performed by the traditional DNNs, and consequently, they are not supported by already existing DNNs hardware accelerators. Moreover, CapsNets require modified data mapping. CapsAcc, a hardware platform targeting CapsNets acceleration, is proposed in~\cite{capsacc} and shown in Fig.~\ref{fig:capsacc}. The computational core of CapsAcc consists of an array of processing elements (PEs), followed by an accumulator that properly adds the partial sums. There is then an activation unit that can apply ReLU, softmax, or squash functions to the output of the accumulator. The activations and weights are stored in data and weight memories, respectively, and there are three buffers used during the computation to exploit data reuse and minimize the accesses to the larger memories. In particular, a data and a weight buffer store the activations and the weights, and a routing buffer is used to store partial results of the dynamic routing iterations. A control unit selects the paths for the mapping of different layers onto the PE array. An efficient on-chip memory management~\cite{Marchisio2019CapStore} can further reduce its energy consumption.

\vspace*{-5pt}

\section{NASCaps: Neural Architecture Search of Convolutional CapsNets}
\label{sec:main_technical}
%@andrea, d by @alberto

Our multi-objective \textit{NASCaps} framework generates and evaluates convolutional- and capsule-based DNNs, by performing a multi-objective NAS, to find a set of accurate yet resource-efficient DNN models, i.e., jointly considering the DNN validation accuracy, energy consumption, latency, and memory footprint. 
The search is based on our specialization of the genetic NSGA-II algorithm~\cite{deb2002fast}, to enable a search with multi-objective comparison and selection among the generated candidate DNNs. 

\begin{figure}[h]
    \centering
    \includegraphics[width=0.95\columnwidth]{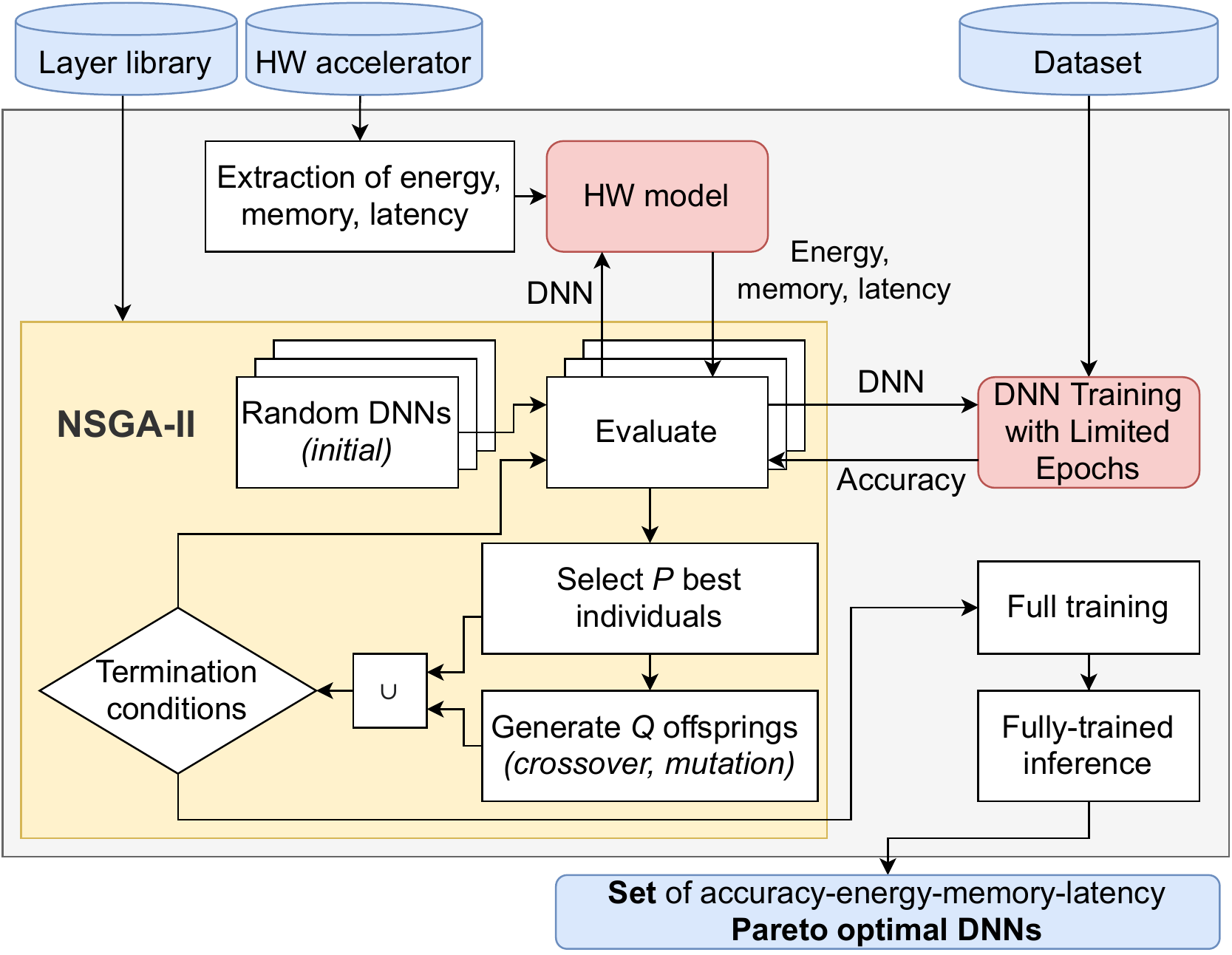} % using drawio 
    \vspace*{-1pt}
    \caption{Overview of our \textit{NASCaps} framework, showing different components and their interconnections defining the workflow.} 
    \label{fig:overall}
    \vspace*{-4pt}
\end{figure}

The overall structure and workflow of the \textit{NASCaps} framework is shown in Fig.~\ref{fig:overall}. As input, it receives the configuration of the underlying hardware accelerator  (that would execute the generated DNN in the real-world scenario) and a given dataset used for DNN training, as well as a collection of the possible types of layers that can be used to form different candidate DNNs. 
First, we create a layer library that includes convolutional layers, capsule layers (as defined in~\cite{sabour}), and the CapsCell and FlatCaps layers defined in~\cite{deepcaps}. We envision that, due to the modular structure of our framework, other types of layers can easily be integrated into its future versions to further extend the search-space, also thanks to the use of a simple modular representation of the candidate networks relying on the combination of single-layer descriptors, as discussed in Section~\ref{subsec:CapsNet_model_SW}. 

The automated search is initialized with $N$ randomly-generated DNNs used as input to start the evolutionary search process. Each candidate DNN is evaluated in terms of its validation accuracy after being trained for a limited number of epochs. As we will discuss in Section~\ref{subsec:reduce_epochs}, this optimization is designed to curtail the computational cost and to reduce the required computational time for the search, while keeping a good level of correlation w.r.t. the full-training accuracy, measured with the Pearson correlation coefficient. 
%training that allows to contain the computational cost and reduce the required computational time (this aspect is detailed and discussed in Section~\ref{subsec:reduce_epochs}). 
Moreover, each DNN under test is also characterized for its energy consumption, latency, and memory footprint, by modeling its inference processing considering the final real-world use case of executing the generated DNN on a specialized DNN hardware accelerator. At this evaluation point, the genetic algorithm proceeds to the next step, finding at each iteration a new Pareto-frontier that contains the best candidate DNN solutions. 
At the end of this selection process, the Pareto-optimal DNN solutions are fully-trained\footnote{A complete training up to the 100\textsuperscript{th} epoch for the MNIST, Fashion-MNIST, and SVHN datasets, and up to the 300\textsuperscript{th} epoch for the CIFAR-10 dataset is conducted.}, to make an exact accuracy evaluation. In the following sub-sections, we discuss the key components of our framework in detail.

\vspace*{-5pt}

\subsection{Parametric Modeling of Capsule Network Layers and Architectures}
\label{subsec:CapsNet_model_SW}
%@andrea, revised by @beatrice

The proposed genetic-based \textit{NASCaps} framework is relying on an explicit position-based representation for each layer of the candidate DNNs. This representation allows to define the key parameters of each layer uniquely.

The DNN layers have been constructed using a \textit{layer descriptor}, which encodes the information needed to build and model a given candidate network, in a very compact form. Each layer descriptor is a 9-element position-based structure, thus guaranteeing the modularity for constructing any given candidate DNN architecture. \textbf{The elements of the layer descriptor are listed as follows:}
\vspace*{-2pt}
\begin{enumerate}
    \item type of layer,
    \item size of the input feature maps $n_{in}$,
    \item number of input channels $ch_{in}$,
    \item number of input capsules $caps_{in}$,
    \item kernel size $kernel_{size}$,
    \item stride size $stride_{size}$,
    \item size of the output feature maps $n_{out}$,
    \item number of output channels $ch_{out}$,
    \item number of output capsules $caps_{out}$.
\end{enumerate}
\vspace*{-2pt}
Such a representation allows to describe even more complex structures by simply defining a new layer \textit{type}. For instance, a layer descriptor can define a more complex repeating structure, composed of multiple elements, like a CapsCell in the DeepCaps architecture.
In this way, the DeepCaps architecture has been described with six layer descriptors. The first one for the single convolutional layer, four CapsCell blocks, and a final Class Capsule layer. 

\begin{figure}[h]
	\centering
	\includegraphics[width=1\linewidth]{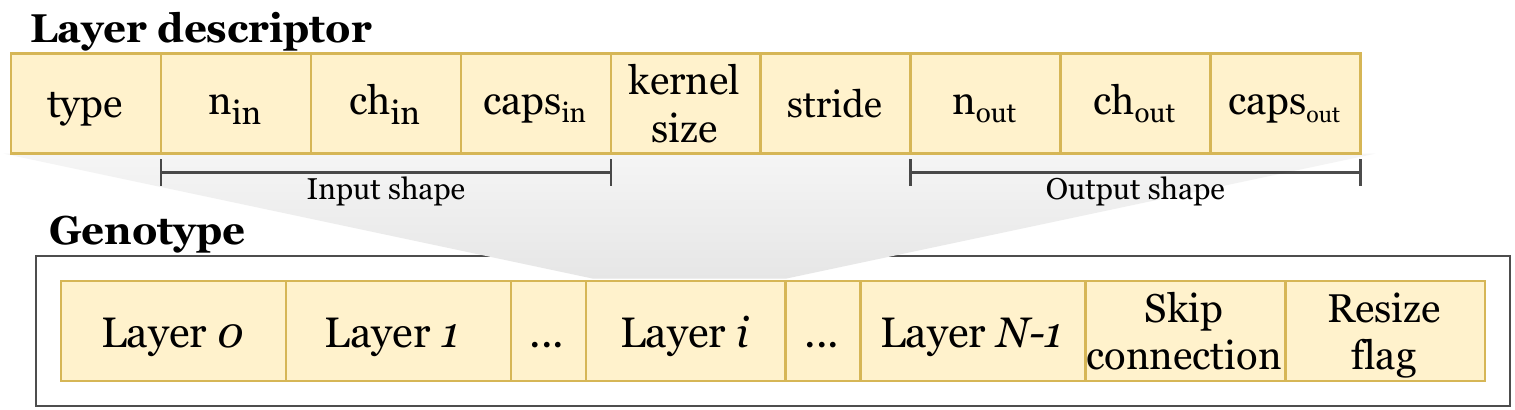}
	\caption{Proposed structure of the genotype.}
	\label{gene}
	\vspace*{-12pt}
\end{figure}

The complete DNN architecture description is then completed by two non-layer terms, that allow to encode the position of a skip connection, and an indicator, called \textit{resize flag}, to explicitly indicate if the resizing of the inputs is required. Fig.~\ref{gene} shows the format proposed to describe a candidate DNN architecture, which is, from now on, referred to as the \textit{genotype}.

\subsection{Modeling the Execution of CapsNets in Hardware Accelerators}
\label{subsec:CapsNet_model_HW}

%@vojta, revisd by @alberto

%\red{support with algorithms and equations for each metric that is modeled}
%\vm{This is updated metrics from the FEECA paper. There are a few changes (accordingly to DATE'19 paper): 16x16 fixed systolic array; 1stage, pewidth=1, 3ns period, membw = 128 (for reading of all columns in one time) }
The \textit{NASCaps} framework can receive any given hardware accelerator executing DNN inference as an input. For illustration, we showcase the modeling of the CapsAcc accelerator; this choice is related to the fact that it supports the execution of all the capsule layers.
Starting from the RTL-level description of the CapsAcc architecture, we extract and model the different micro-architectural configurations at a higher abstraction level, which constitutes the inputs for our model. First, the description of the operation-specific parameters of the layers is presented. Afterward, the global parameters, that are strictly related to the CapsAcc accelerator, are discussed.

\vspace*{-2pt}

\subsubsection{Operation-Specific Modeling for different Layers}\ \\
The operation-specific parameters that can be extracted from the execution of different operations in the hardware are the following:

\vspace*{-2pt}
\begin{itemize}[leftmargin=*]
    \item $weights$: number of weights in the layer,
    \item $sums\_per\_out$: number of terms to be added for an output value,
    \item $data\_per\_weight$: number of feature maps that are multiplied by the same weight.
\end{itemize}
\vspace*{-2pt}
For each operation, these parameters are computed by different equations, due to the different nature of the respective types of computations, see Table~\ref{tab:equation_layers}. Note that, by setting $caps_{in}$ and $caps_{out}$ to $1$, the \textit{ConvCaps} and \textit{ClassCaps} layers become a traditional convolutional layer and fully-connected layer, respectively.

\begin{table*}[h]
\caption{Equations for the operation-specific modeling of CapsNets.} 
\vspace*{-2pt}
\label{tab:equation_layers}
\resizebox{.8\textwidth}{!}{
\begin{tabular}{l|c|c|c}
\toprule
 \multicolumn{1}{c}{\textbf{Operation}} &
  \multicolumn{1}{c}{\textbf{$weights$}} &
  \multicolumn{1}{c}{\textbf{$sums\_per\_out$}} &
  \multicolumn{1}{c}{\textbf{$data\_per\_weight$}} \\ \midrule
\textbf{ConvCaps layer}    & $(ch_{in} \cdot kernel_{size}^2+1) \cdot ch_{out} \cdot caps_{out} \cdot caps_{in}$ & $(kernel_{size}^2+1) \cdot ch_{in} \cdot caps_{in}$ & $(n_{out})^2 \cdot ch_{in} \cdot caps_{in}$ \\
\textbf{ConvCaps3D layer}   & $(ch_{in} \cdot kernel_{size}^3+1) \cdot ch_{out} \cdot caps_{out} \cdot caps_{in}$ & $(kernel_{size}^3+1) \cdot ch_{in} \cdot caps_{in}$ & $(n_{out})^2 \cdot ch_{in} \cdot caps_{in}$   \\
\textbf{ClassCaps layer}     & $(ch_{in} \cdot n_{in}^2+1) \cdot ch_{out} \cdot caps_{out} \cdot caps_{in}$ & $(n_{in}^2+1) \cdot ch_{in} \cdot caps_{in}$  & $1$       \\
\textbf{Dynamic Routing} & $ch_{in} \cdot kernel_{size}^2 \cdot ch_{out}$ & $caps_{in}$  & $1$ \\ \bottomrule
\end{tabular}}
\vspace*{-2pt}
\end{table*}

\vspace*{-2pt}

\subsubsection{Global Parameter Modeling}\ \\
Our models estimate the latency and the energy consumption of the inference of one input, for a given \textit{CapsNet}, while the memory footprint is computed as the sum of the number of weights for each layer. They are modeled for each operation in a modular way (i.e., bottom-up). First, the weights must be loaded onto the PE array, then reused as long as they need to be multiplied by other inputs. Afterward, the next group of weights is loaded until all the computations of the layers are done (see Eqs.~\ref{eq:wloadcycles}-\ref{eq:cycles}). 
The model has been validated by comparing the results with the hardware implementation of the CapsAcc~\cite{capsacc} accelerator. 
The adopted model parameters are the following:

\begin{itemize}[leftmargin=*]
    \item $w\_load\_cycles$: number of clock cycles required to load the weight onto the PE array,
    \item $w\_loads$: number of groups of weights loaded onto the PE array,
    \item $cycles(l)$: number of cycles required to execute the layer $l$,
    \item $ma$: number of memory accesses,
    \item $en_{mem}$: energy consumption of a single memory accesses,
    \item $pwr_{PEA}$: power consumption of the PE array.
\end{itemize}

\vspace*{-10pt}
%\subsubsection{Memory operations}
\begin{equation}
\label{eq:wloadcycles}
w\_load\_cycles = 16
\end{equation}
\begin{equation}
\label{eq:wloads}
w\_loads = \left\lceil { weights \over { 16  \cdot  \min{(16, sums\_per\_out)}} } \right\rceil 
\end{equation}
%\subsubsection{Number of cycles}
\begin{equation}
\label{eq:cycles}
cycles(l) = w\_load\_cycles \cdot w\_loads+data\_per\_weight
\end{equation}

%$$pwr(l) = n_{rows}\cdot n_{cols} \cdot pwr_{pe}(n_{pe}, T) +  pwr_{mem}(mem_{bw})$$

The overall latency is then computed as the sum of the contributions of the layers (Eq.~\ref{eq:latency}).
\begin{equation}
\label{eq:latency}
latency = \sum_{l \in L}{cycles(l) \cdot T}
\end{equation}

In the Eq.~\ref{eq:memory}, the number of memory accesses is computed by distinguishing whether the operation is a convolutional layer or not. Such a distinction has been implemented by analyzing the value of $data\_per\_weight$, which is greater than 1 for convolutional layers and 1 otherwise.
\begin{equation}
\label{eq:memory}
ma = \begin{cases}
    256,\ \ \ \ \ \ \ \ \ \ \ \ \ \ \ \ \ \ \ \ \ \ \ \ \ \text{if } data\_per\_weight = 1  \\ % 16x16
    16 \cdot \max(sums\_per\_out - 15, 1),\ \text{otherwise}
    \end{cases}
\end{equation}

The energy of the accelerator (Eq.~\ref{eq:energy}) is estimated as the sum of the energy of memory accesses and the sum of the power consumption of each layer processed in the PE array, multiplied by its latency (period $T$ and the number of cycles). Note that the average power consumption of the PE array is used in our model. 
%$$\resizebox{\linewidth}{!}{
\begin{equation}
\label{eq:energy}
energy = \left\lceil \frac{ma \cdot 8}{128}  \right\rceil \cdot en_{mem} + \sum_{l \in L}{cycles(l) \cdot T \cdot pwr_{PEA}}
\end{equation}
%}$$

%The overall area on a chip of the PE array of the CapsAcc accelerator does not depend on the CapsNet but only on the hardware parameters.% size of the systolic array in particular.
%$$area = n_{rows}\cdot n_{cols} \cdot area_{pe}(n_{pe}, T) + area_{mem}(mem_{bw})$$

\vspace*{-8pt}

\subsection{The Multi-Objective NSGA-II Algorithm}
\label{subsec:genetic algorithm}
%@andrea, revised by @vojta and @alberto

The selection of the Pareto-optimal solutions for the \textit{NASCaps} framework is based on the evolutionary algorithm NSGA-II~\cite{deb2002fast}. It has a main loop (lines 2-14 of Algorithm~\ref{alg:nsga}) whose iterations represent a single generation of the overall evolution process of an initial population. The initial population (sized \textit{n}) is randomly generated and can be referred to as $ P_1 $ (line 1 of Algorithm~\ref{alg:nsga}). This set of solutions represents the initial parent generation of the algorithm. 
The crossover among the solutions belonging to $ P_t $ (line 3) allows the generation of a new set of offspring individuals $ Q_t $. At this point, the population $P_t \cup Q_t$ is sorted according to a non-domination criterion. For each iteration of the inner loop (lines 6-13), the candidate solutions are grouped into different fronts \textbf{$ F_i $}. The ones included in the first front \textbf{$ F_1 $} represent the best-found solutions of the overall population. Each subsequent front ($F_2, F_3, \dots$) is instead constructed by removing all the preceding fronts from the population and then finding a new Pareto-front. Since the first front may be composed of less than \textit{n} individuals, also the solutions from subsequent fronts will be selected to be part of the next parent generation. 

\begin{figure}[h]
	\centering
	\vspace*{1pt}
	\includegraphics[width=1\linewidth]{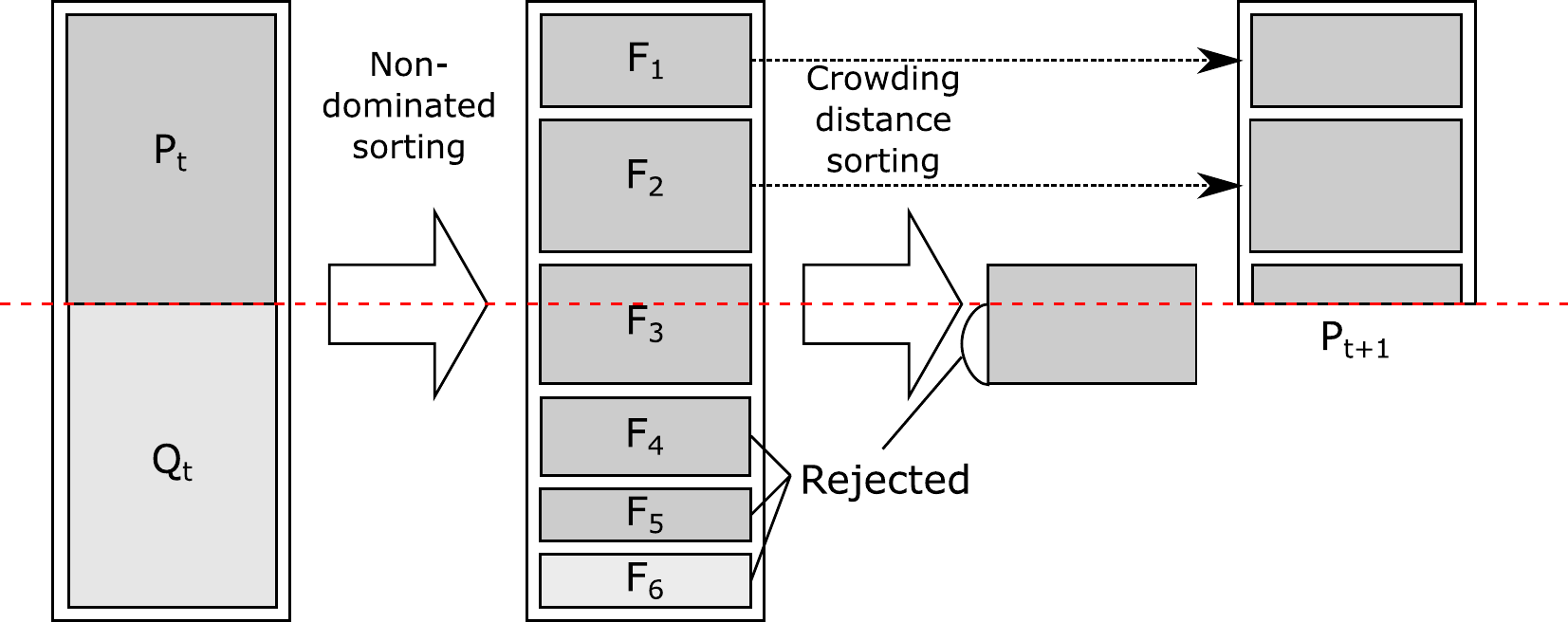}
	\caption{Sorting of the population.}
	\label{nsgaii}
	\vspace*{-6pt}
\end{figure}

To have exactly \textit{n} parents in the output set, the solutions that are part of the last front are ranked using the crowded distance comparison approach (line 11), which consists of sorting the population of that front according to each objective function value in ascending order. These steps are shown in Fig.~\ref{nsgaii}.
Only half of the population becomes part of the next parent generation, while the other half is discarded. 

These steps repeat for a certain number $g$ of generations. The complete pseudocode is reported in Algorithm~\ref{alg:nsga}, where the following procedures are used: 
\vspace*{-2pt}
\begin{itemize}[leftmargin=*]
	\item $RandomConfigurations(n)$ randomly generates $n$ configurations belonging to the search space.
	\item $CrossoverAndMutate(X, n)$ generates $n$ new offsprings from parents $P$ by crossover and mutation (described in Section \ref{sec:crossover}).
	\item $EstimateParameters(X)$ evaluates the new candidate solutions from a set $X$.
	\item $PickPareto(X)$ selects the Pareto-optimal solutions from a set $X$, and these solutions are removed from the set.
	\item $DistanceCrowding(X, n)$ returns $n$ solutions from a set $X$ (as described in \cite{deb2002fast}).
\end{itemize}
\vspace*{-2pt}

\begin{figure}[h]
\vspace*{-2pt}
	\begin{algorithm}[H]
	\captionsetup{font=small}
		\caption{\textbf{: The genetic NSGA-II algorithm used in our \textit{NASCaps} framework.}}
		\label{alg:nsga}
		\vspace*{-2pt}
		\input{nsga.tex}
		\vspace*{-2pt}
	\end{algorithm}
	\vspace*{-10pt}
\end{figure}

The advantage of a multi-objective algorithm lies in the fact that it re-constructs the Pareto-front at each generation, aiming to cover all the possible solutions. The algorithm's output is a set of non-dominated solutions.

\vspace*{-2pt}

\subsubsection{Crossover and mutation operations}\label{sec:crossover}\ \\
The two key operators in the progression of a genetic algorithm are crossover and mutation.
The standard single-point \textbf{crossover} operation allows to generate the offspring solutions, given two parent solutions $ P_a $ and $P_b$ that have been previously randomly picked among the current population candidates. The genotypes of the two parent individuals are split into two parts each. The splitting point is pseudo-randomly selected. Initially, a cut point is randomly chosen. Then, a series of checks are performed to verify the validity of the output genotypes.
The following criteria have been applied to choose the splitting point correctly:
\vspace*{-2pt}
\begin{itemize}[leftmargin=*]
	\item the cut-points are chosen to ensure that the generated DNN is made up of at least one initial convolutional layer and a minimum of 2 capsule layers,
	\item no convolutional layer is placed between two capsule layers. 
\end{itemize}

\begin{figure}[h]
	\centering
	\includegraphics[width=1\linewidth]{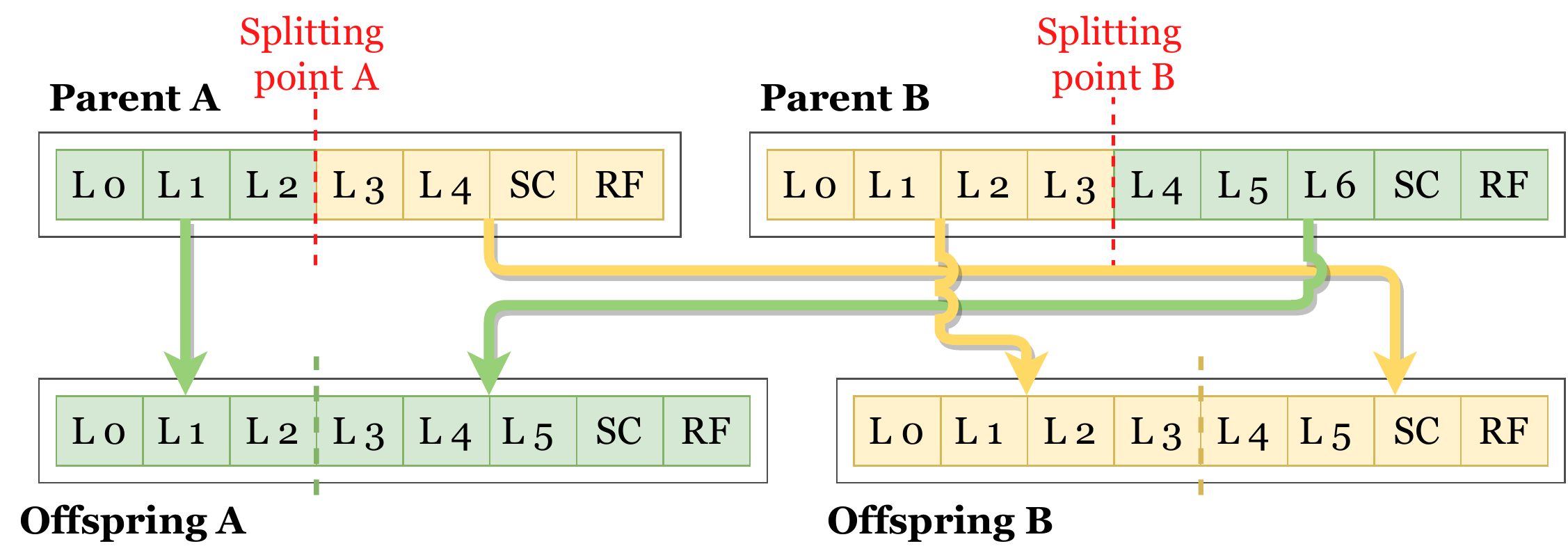}
	\caption{Example of crossover between two genotypes.}
	\label{crossover}
	\vspace*{-5pt}
\end{figure}

Note, the reason behind the second constraint is that capsules aim to derive higher-level information w.r.t. convolutional layers. At this point, the actual crossover operation is performed. As shown in Fig.~\ref{crossover}, the last parts of the parent genes $ P_a $ and $ P_b $ are switched.

The second key operation performed by the algorithm is \textbf{mutation}. As it has been implemented for our \textit{NASCaps} framework, the operator performs a mutation by randomly choosing one of the layer descriptors from the genotype of the input candidate network, and by randomly modifying one of the main parameters of the selected layer with a probability $p_m$.
In particular, the parameters that can be affected by a mutation are the kernel size, the strides, the number of output capsules, and the position of the skip connection.

After these two operations, a further step is performed to ensure the validity of the output genotypes that, in a large number of cases, will represent an invalid DNN. This correction step allows to properly adjust the input and output tensor dimensions for every layer for genotypes derived from a mutation or a crossover operation, which can randomly modify or join different parent genotypes together.

\vspace*{-5pt}

\section{Evaluating our NASCaps Framework}
\label{sec:results}

\vspace*{-2pt}

\subsection{Experimental Setup}
\label{subsec:exp_setup}

%@vojta, revised by @andrea

\begin{figure}[h]
	\centering
	\includegraphics[width=1\linewidth]{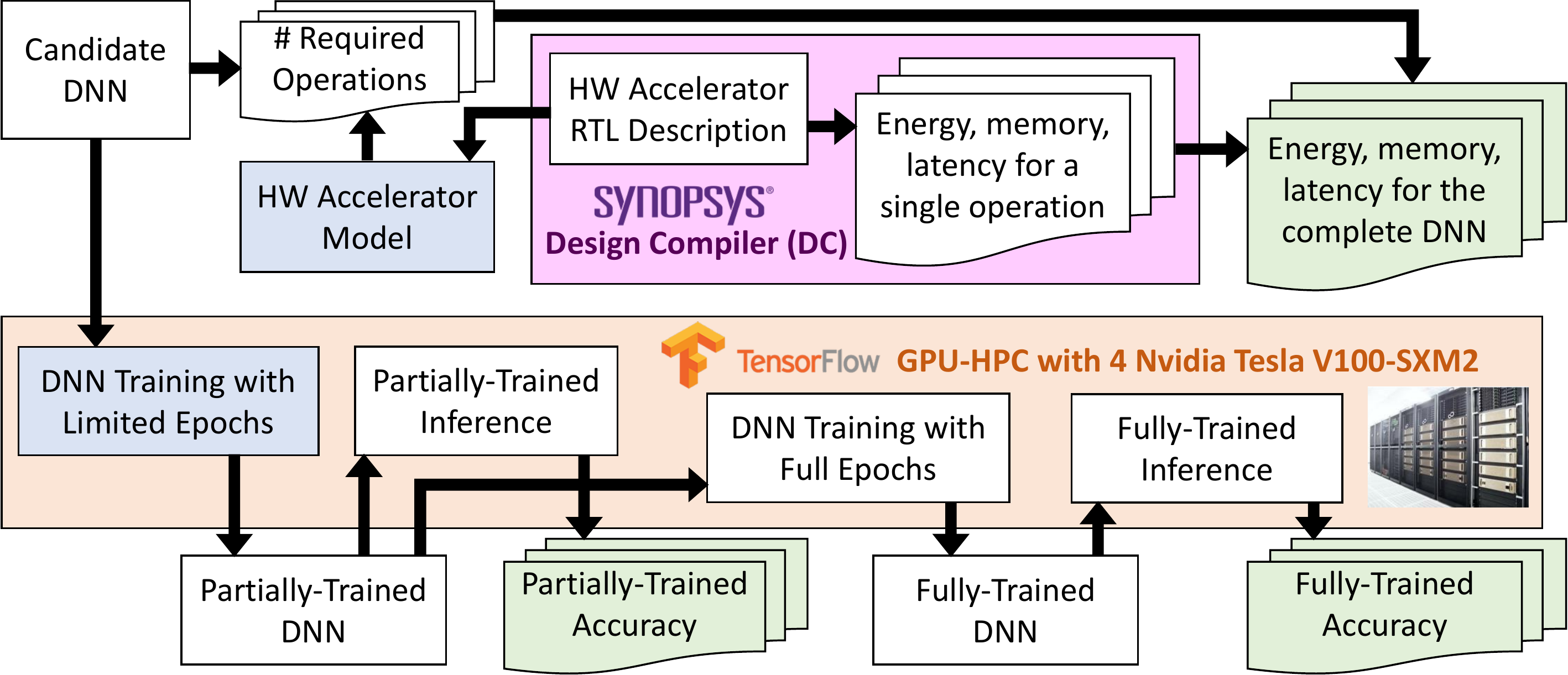}
	\caption{Setup and tool-flow for conducting our experiments.}
	\label{Exp_setup}
	\vspace*{-5pt}
\end{figure}

%\red{figure with exp setup and toolflow}
%\vm{Our overall image Fig~\ref{fig:overall} is so general that we don't need to draw another figure with the same information. And we do not have a space for that} \red{Alberto: I think a small figure can be done (I can do it). The space is not an issue because we can reduce the font size up to 9pt}
The overview of our experimental setup and tool-flow is shown in Fig. \ref{Exp_setup}.
The training and testing for accuracy of the candidate DNNs have been conducted with the TensorFlow library~\cite{tensorflow}, while extensive experiments are performed using the GPU-HPC computing nodes equipped with four NVIDIA Tesla V100-SXM2 GPUs. Our proposed \textit{NASCaps} framework has been evaluated for the MNIST~\cite{MNIST}, Fashion MNIST (FMNIST)~\cite{Fashion-MNIST}, SVHN~\cite{SVHN} and CIFAR-10~\cite{CIFAR} datasets.
The implementation of the HW model is based on an open-source SoA Capsule Network  accelerator (CapsAcc)~\cite{capsacc}, as described in Section~\ref{subsec:CapsNet_model_HW}. The core processing elements were synthesized using the Synopsys Design Compiler with a 45nm technology node and a clock period $T$ of 3ns. 

The experiments were divided into three steps. (1) In the first step, a basic random search has been performed, to investigate how many training epochs are necessary to train the candidate DNNs and evaluate their accuracy in the loop of the genetic NSGA-II algorithm. (2) During the second step, the search algorithm for finding Pareto-optimal DNN architectures for the energy, memory, latency, and accuracy objectives is executed. (3) Finally, the selected Pareto-optimal DNNs have been fully-trained. To evaluate the transferability of the selected DNNs w.r.t. different datasets, the selected DNNs have been fully-trained also for the other datasets. 
Moreover, the following settings have been used to conduct the experiments:
\vspace*{-2pt}
\begin{itemize}[leftmargin=*]
\item Initial parent population size $|P|= 10$
\item Offspring population size: $|Q|=10$
\item Maximum number of generations for the genetic loop: $g=20$
\item Mutation probability: $p_m=10\%$
\item $kernel_{size} \in \{3\times 3, 5 \times 5, 9 \times 9 \}$
\item $stride_{size} \in \{1, 2\}$
\item $ch_{out} \in \{1, 2, \dots , 64\}$
\item $caps_{out} \in \{1, 2, \dots, 64\}$
%\item Skip connections: pseudo-randomly chosen to avoid having extra layers in parallel with the main sequence of layers (i.e., the LDs), to keep a simple yet readable genotype description of the candidate networks.
\end{itemize}
%The initial parent population size was 10, as well the offspring one and the total number of generations for the genetic loop has been setted to a maximum of 20. %\blue{Andrea: Is this value ok? I guess not because of the results' images. I think that the number of generation should be the same overall} \red{Alberto: I agree: either the same for every dataset, or we don't specify anything here and we report the value in the results specific for each dataset.} 
%Mutation occured with a probability of 10\%. The kernel size is randomly chosen among common values 3x3, 5x5, 9x9 and, for the fully connected layer, the kernel size is equal to the input feature map size. Strides are instead chosen between 1 and 2. The number of output channels and capsules is randomly chosen with a maximum value of 64. The skip connection has been instead pseudo-randomly chosen in order to have no need of extra layers in parallel with the main sequence of layers (the ones defined through LDs) in order to keep a simple readable genotype description of the candidate network. 

These values, as well as the training hyper-parameters (e.g., batch sizes, number of epochs and learning rate) have been selected by conducting a set of preliminary experiments and considering reasonable runtimes. 

\vspace*{-7pt}

\subsection{Results for Reduced Training Epochs for Full-Training Accuracy Estimation}
\label{subsec:reduce_epochs}
%@andrea, revised by @vojta
\vspace*{-2pt}

One of the most crucial aspects linked to the NAS problem lies in its high computational exploration cost. This is due to the \textit{large number of candidate networks} that constitute the population and the \textit{time-consuming training steps needed to evaluate the accuracy.} 
To limit the time needed to perform the complete search and consequently, its computational cost, we propose a two-stage evaluation approach. (1) The first step consists of training the population of candidate networks with a limited number of epochs, producing a set of partially-trained DNNs. The validation accuracy obtained by the partially-trained DNNs has been used for the evaluation of the Pareto-fronts in the NSGA-II algorithm, as discussed in Section~\ref{subsec:genetic algorithm}. 
The choice of the number of epochs has been determined carefully by analyzing the impact of different epoch sizes over the achieved accuracy for different datasets. (2) Afterward, the candidate networks that show their accuracy and hardware efficiency in a Pareto-front are fully-trained to evaluate their actual validation accuracy.

Hence, this approach allowed to use only a reduced number of training epochs to predict the full-training accuracy of the DNNs. \textit{This approach has been tested using 66 randomly generated DNNs (in addition to CapsNet and DeepCaps architectures) and performing a full-training on them}, while recording the obtained validation accuracy at each training epoch. The Pearson correlation coefficient ($PCC$)~\cite{PearsonCoefficient} has been computed to analyze the correlation between the accuracy of the fully-trained DNNs and the accuracy of the same DNNs at the intermediate steps.

Table~\ref{tab:pcc} shows the values of the $PCC$, computed between the accuracy of the DNNs after $n$ training epochs and their accuracy after a full training. The median cumulative training time needed to perform an $n$ epoch training is also reported. 
This study allowed to determine, as expected, that more complex datasets require a larger number of training epochs to distinguish the most promising networks from the rest correctly.
For the case of the MNIST dataset, 5 epochs are sufficient to reach a $PPC$ equal to 0.9999. Instead, for the CIFAR-10 dataset, such a high value of confidence is never reached within the first few epochs. In this case, 10 training epochs are selected, which ensure a $PCC$ equal to 0.9334. This choice leveraged the tradeoff between the correlation coefficient and the required training time. Of course, a larger number of training epochs can also be selected, but it would drastically increase the exploration time due to the DNN training, which is a crucial parameter to consider when large populations and/or number of generations are explored by the \textit{NASCaps} framework. On the other hand, the selection performed after 10 epochs of training allowed to discard more Pareto-dominated candidate networks than what would have been discarded after 5 epochs.
For the Fashion-MNIST and SVHN datasets, the selection stage has also been performed after 5 epochs of training. 
%\vm{We used a following settings - (max epochs; max training time; total wall time): MNIST: (5, 5 minutes, 12 hours), FMNIST (same as MNIST), SVHN (5, 10 minutes, 24 hours), CIFAR-10 (10, 10 minutes, 24 hours)}

\begin{table}[h]
\vspace*{-4pt}
\captionsetup{font=small}
	\caption{Pearson correlation coefficient (PCC) and median cumulative training time expressed in seconds (MCTT) for the MNIST, Fashion-MNIST (FMNIST), SVHN and CIFAR-10 datasets.}
	\label{tab:pcc}
	\vspace*{-3pt}
	\resizebox{.95\linewidth}{!}{
    \begin{tabular}{lccccccc} 
\hline
\multicolumn{2}{l}{ \textbf{Epoch n.} }                              & \textbf{1}      & \textbf{3}           & \textbf{5}           & \textbf{10}          & \textbf{15}          & \textbf{20}           \\ 
\hline
\multirow{2}{*}{ \textbf{MNIST} }        & PCC                       & 0.8407 & 0.9998               & 0.9999               & 1.0000               & 1.0000               & 1.0000                \\
                                         & \multicolumn{1}{l}{MCTT} &  55.4  & 166.2 & 277.0  & 554.0  & 831.0 & 1108.0  \\ 
\hline
\multirow{2}{*}{\textbf{FMNIST} } & PCC                       & 0.8306 & 0.8963               & 0.9013               & 0.9935               & 0.9989               & 0.9998                \\
                                         & \multicolumn{1}{l}{MCTT} & 86.2       & 258.7 & 431.1 & 862.3 & 1293.4 & 1724.6  \\ 
\hline
\multirow{2}{*}{\textbf{SVHN} }          & PCC                       & 0.6812 & 0.8733               & 0.9518               & 0.9531               & 0.9667               & 0.9876                \\
                                         & \multicolumn{1}{l}{MCTT} &  128.3      & 385.0 & 641.6 & 1283.3 & 1924.9 & 2666.6  \\ 
\hline
\multirow{2}{*}{\textbf{CIFAR-10} }      & PCC                       & 0.2969 & 0.4259               & 0.7279               & 0.9334               & 0.9518               & 0.9879                \\
                                         & \multicolumn{1}{l}{MCTT} &  61.6      & 184.7 & 307.9 & 615.8 & 923.6 & 1231.5  \\
\hline
\end{tabular}
    }
    \vspace*{-3pt}
\end{table}

Note, a certain set of networks can be discarded relatively early, i.e., after a few training epochs, since they do not improve their accuracy much. 
The candidate networks that pass the selection stage can then complete their training. A second selection stage is beneficial for performing a more fine-grained selection of the candidate networks, and avoiding the tedious and computational-hungry full-training of Pareto-dominated DNNs.

\vspace*{-5pt}

\subsection{NASCaps Results for the Partially-Trained DNNs}
%@andrea, revised by @vojta

Our \textit{NASCaps} framework is first applied to the MNIST dataset to evaluate its efficiency and correct behavior. The number of generations is set at 20, but a maximum time-out of 12 hours has been imposed in the cases of the MNIST and Fashion-MNIST datasets, while a 24-hour maximum search time has been used for the CIFAR-10 and SVHN datasets.

\begin{figure}[h] %moved to bottom for better formatting
\vspace*{-2pt}
    \begin{subfigure}{\columnwidth}
        \centering
        \includegraphics[width=0.95\columnwidth]{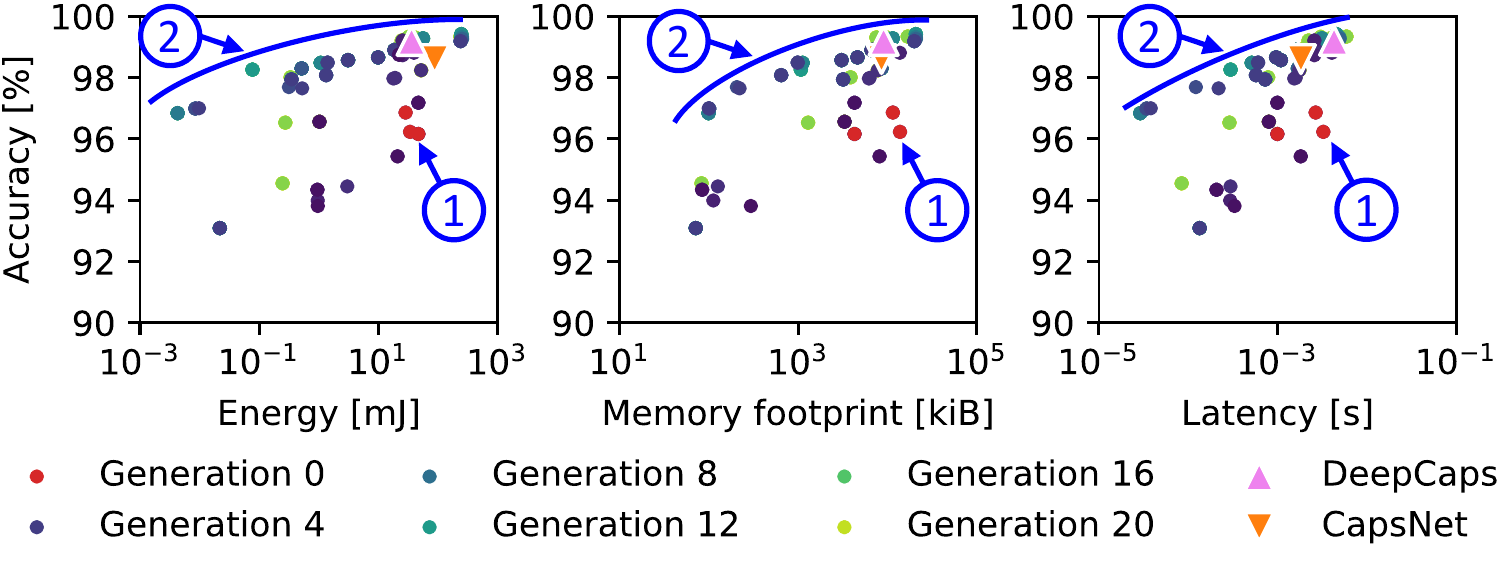}
        \vspace*{-4mm}
        \caption{}
        \label{fig:res_nas_mist}
    \end{subfigure} \\
    \begin{subfigure}{\columnwidth}
        \centering
        \includegraphics[width=0.95\columnwidth]{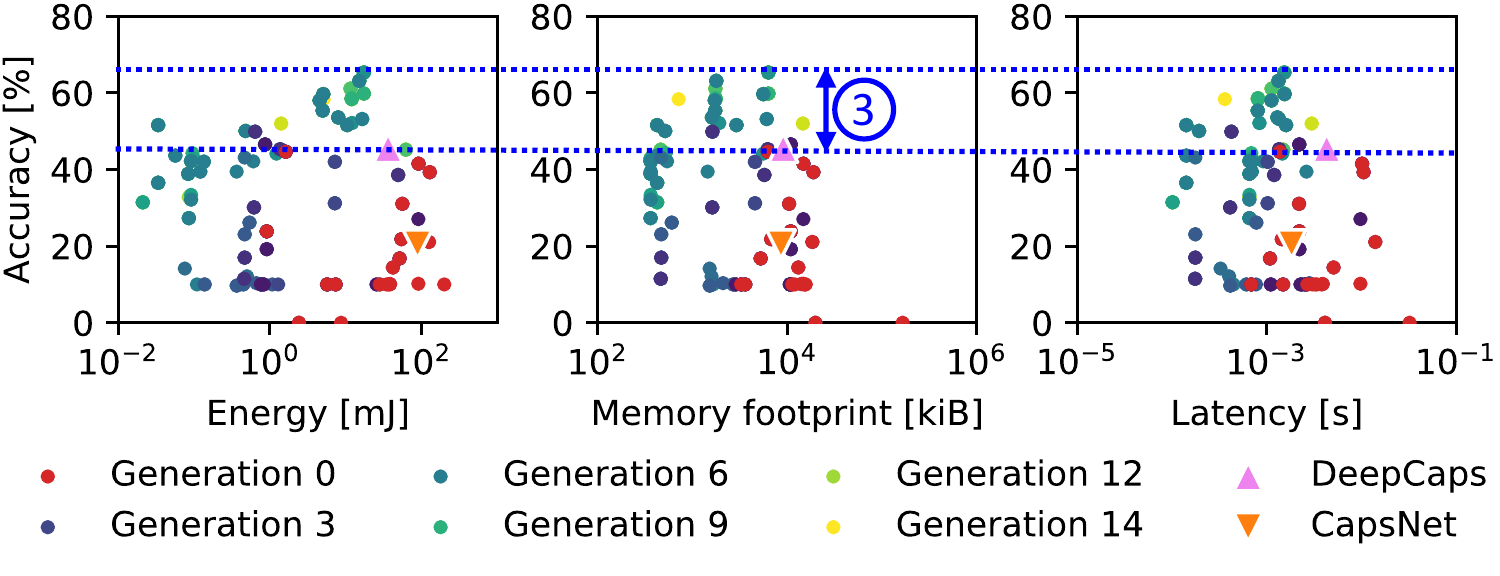}
        \vspace*{-4mm}
        \caption{}\label{fig:res_nas_cifar}
    \end{subfigure}
    \vspace*{-5pt}
    \caption{Partially-Trained DNN NAS for (a) the MNIST dataset, and (b) the CIFAR-10 dataset. The color shows in which generation the solution occurs first. 
    %\red{The X for DeepCaps and CapsNet sometimes are not visible. They should be displayed after the other points to appear in the foreground and also make them bigger}}% \vm{in the img folder FMNIST is included as well}}
    }
    \label{fig:res:nas}
    \vspace*{-3pt}
\end{figure}

The search for the \textit{MNIST-NAS} lasted for 20 complete generations, and the single candidate networks were trained for 5 epochs. 
This setup led to train and evaluate a total of 210 DNNs. The resulting individual solutions are compared to the two reference SoA solutions, that are the CapsNet and DeepCaps architectures. In Fig.~\ref{fig:res_nas_mist}, each individual DNN architecture is represented w.r.t. the four objectives of the search. 

The Fashion-MNIST search ended at its 19\textsuperscript{th} generation (in 12 hours) and evaluated a total of 200 candidate architectures.
The search for the SVHN dataset lasted for 12 generations, and it allowed to evaluate 130 architectures. 
For the CIFAR-10 dataset, the search reached its 14\textsuperscript{th} generation, with a total of 150 tested architectures.

Fig.~\ref{fig:res:nas} shows how the evolutionary search algorithm progressed for the MNIST and CIFAR-10 datasets. 
Note that the red dots, i.e., the initial population at the generation 0 (see pointer \rpoint{1} in Fig.~\ref{fig:res_nas_mist}), represent \textit{randomly generated} DNNs. The objectives significantly improve during the following iterations (see pointer \rpoint{2}), when our evolutionary algorithm finds better candidate DNN architectures using crossover and mutation operations iteratively.
The reduced epoch training allowed to evaluate a large number of candidate networks (a total of nearly 700 architectures) based on convolutional and capsule layers. This method led to finding multiple candidate architectures that have been able to reach an accuracy up to 30.86\% higher than the best among the partially-trained SoA solutions, i.e., within the limits of a strongly reduced training time. For instance, the NAS for the CIFAR-10 dataset produced a network with an accuracy of 76.46\% after 10 epochs, while the DeepCaps architecture reached only 45.60\% accuracy within the same training interval (see pointer \rpoint{3} in Fig.~\ref{fig:res_nas_cifar}). This corroborates the fact that our \textit{NASCaps} can generate networks with higher accuracy compared to DeepCaps-like structures, when %both are subjected
constrained to short training time.% constraints.

%Conclusions from Fig~\ref{fig:res:nsa}
%\begin{itemize}[leftmargin=*]
%    \item For MNIST only a few generations are needed
%    \item For complex CIFAR-10 dataset the best solutions were discovered later
 %   \item We are able to get after 10 (or 5) epochs better accuracies than SoA!
%\end{itemize}

%Conclusions from Fig~\ref{fig:res:final}
%\begin{itemize}[leftmargin=*]
%    \item Random search is not working ... ufff! :) the majority of the solutions was out of the scope
%    \item the solutions found for the CIFAR-10 are too complex for the MNIST
%    \item the solutions from MNIST are not working on the more complex dataset as good as the solutions optimized for the dataset in particular.
%    \item For MNIST we got better results than SoA, for CIFAR-10 we beated CapsNet, however the DeepCaps are slightly better. But We got significantly better hw parameters.
%\end{itemize}

\vspace*{-5pt}

\subsection{NASCaps Results for the Selected Fully-Trained DNNs}
\label{subsec:NASCaps_results}
%@andrea, revised by @vojta

\begin{figure}[h]
    \begin{subfigure}{\columnwidth}
        \centering
        \includegraphics[width=0.95\columnwidth]{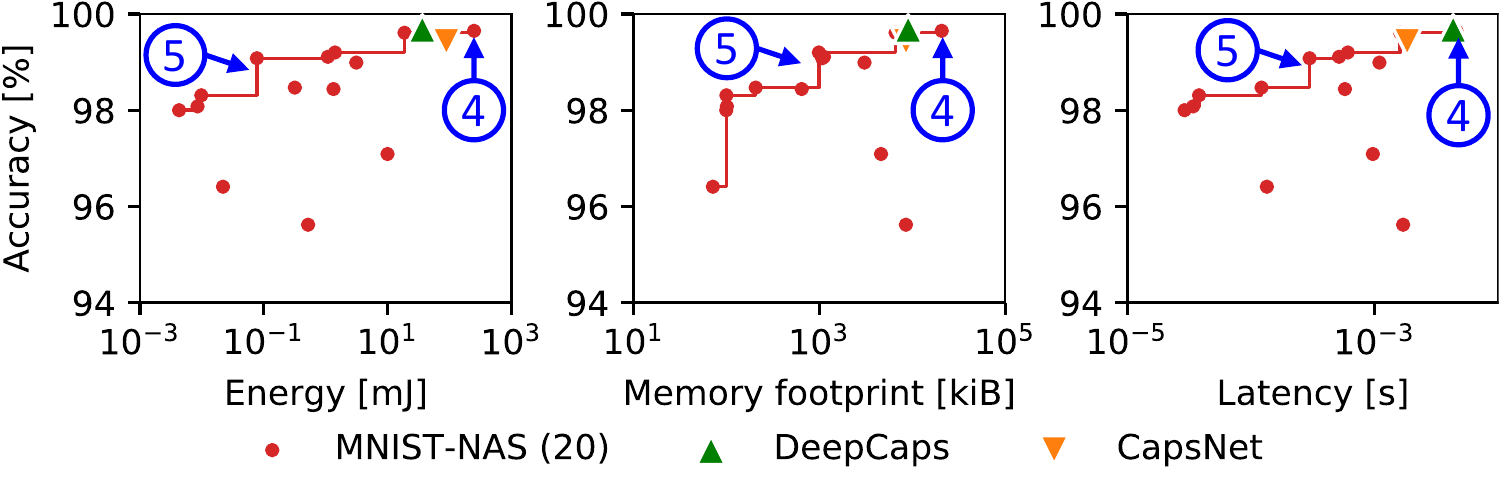}
        \vspace*{-4mm}
        \caption{}\label{fig:res_final_mnist}
    \end{subfigure}
    \begin{subfigure}{\columnwidth}
        \centering
        \includegraphics[width=0.95\columnwidth]{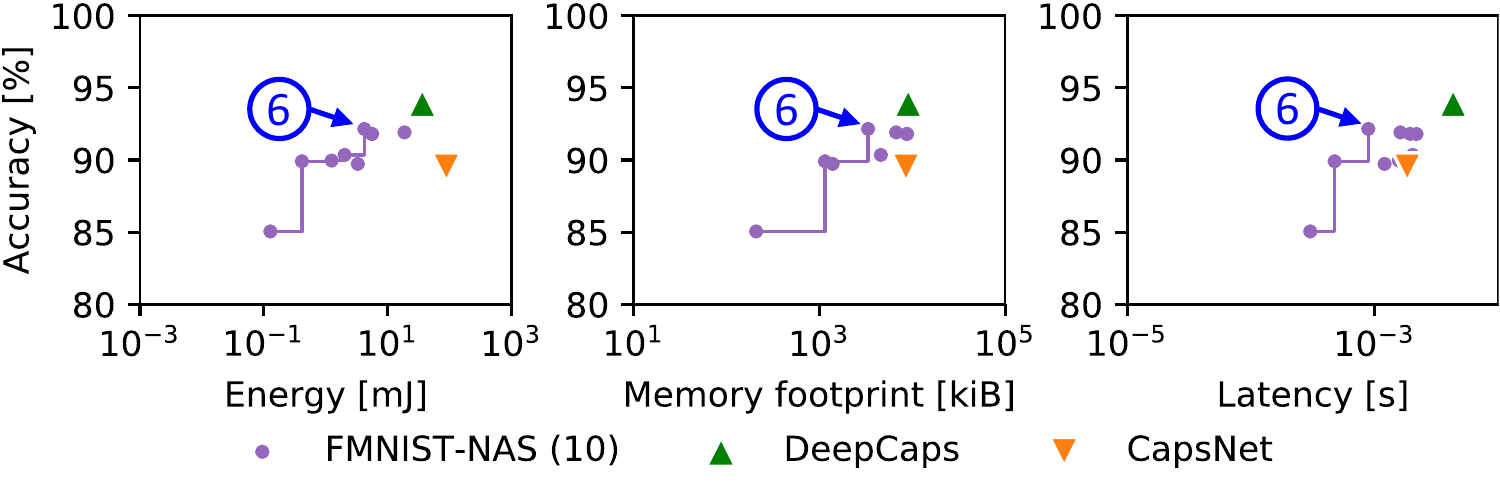}
        \vspace*{-4mm}
        \caption{}\label{fig:res_final_fmnist}
    \end{subfigure}
    \begin{subfigure}{\columnwidth}
        \centering
        \includegraphics[width=0.95\columnwidth]{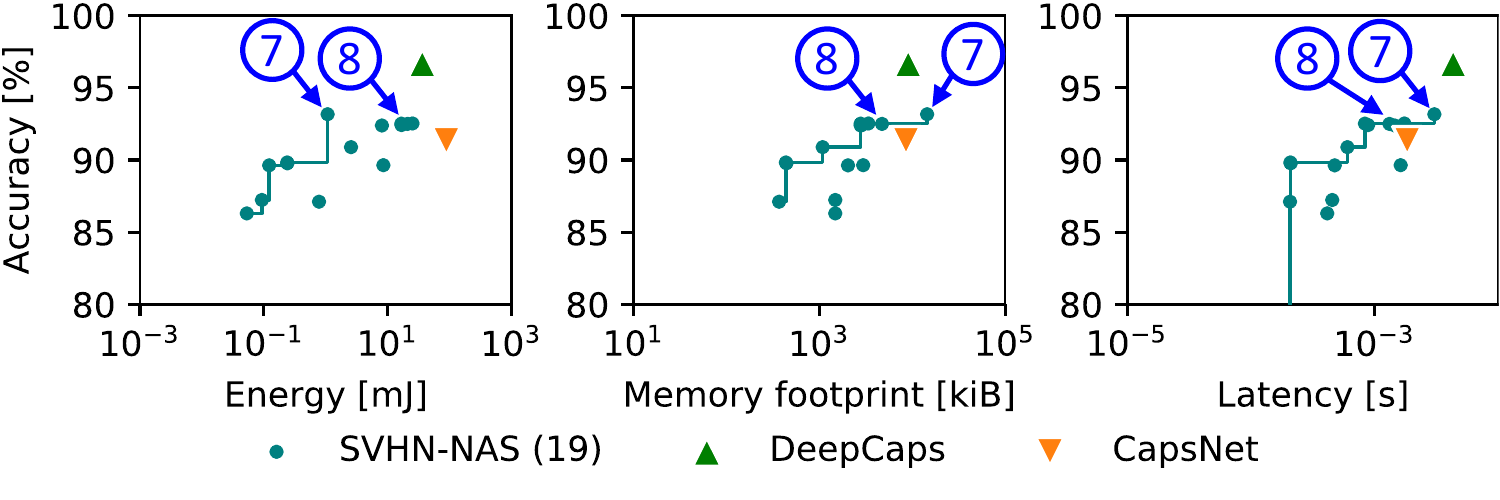}
        \vspace*{-4mm}
        \caption{}\label{fig:res_final_svhn}
    \end{subfigure}
    \begin{subfigure}{\columnwidth}
        \centering
        \includegraphics[width=0.95\columnwidth]{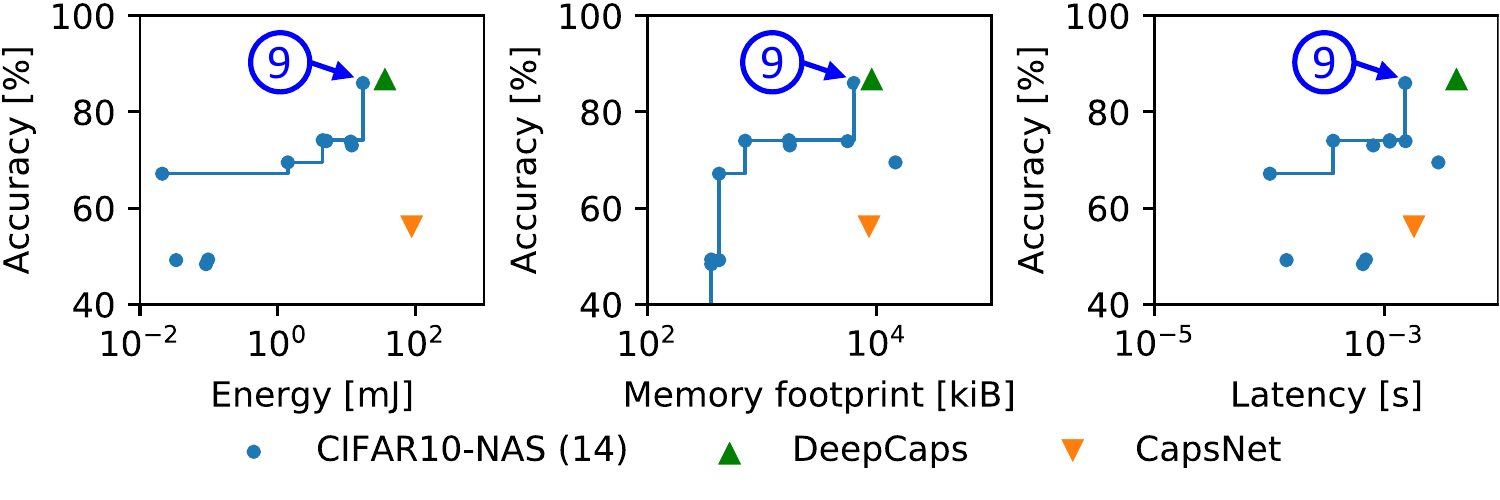}
        \vspace*{-4mm}
        \caption{}\label{fig:res_final_cifar}
    \end{subfigure}
    \vspace*{-5pt}
    \caption{Fully-trained DNN results for (a) the MNIST, (b) the Fashion-MNIST, (c) the SVHN, and (d) CIFAR-10 datasets. }
    \label{fig:res:final}
    \vspace*{-8pt}
\end{figure}

After the first selection stage, the candidate DNNs belonging to the Pareto-optimal subsets have been fully-trained to evaluate their final accuracy. Fig.~\ref{fig:res:final} shows the Pareto-optimal solutions at the end of the full-training process.

\subsubsection{NASCaps for the MNIST Dataset}\ \\
The highest-accuracy architecture (see pointer \rpoint{4} in Fig.~\ref{fig:res_final_mnist}) found during the MNIST search allowed to reach an accuracy of 99.65\% in 93 epochs of training. However, that particular solution requires 2.8$\times$ more energy, 2.5$\times$ more time, and 2.4$\times$ more memory w.r.t. the CapsNet architecture. 
The red front (see pointer \rpoint{5}) also highlights other interesting solutions belonging to the derived Pareto-optimal front, with a slightly lower accuracy, but \textit{up to a couple of orders of magnitude lower energy, memory and latency} achieved by our identified solutions. 

\vspace*{-4pt}
\subsubsection{NASCaps for the Fashion-MNIST Dataset}\ \\
One of the best solutions (see pointer \rpoint{6} in Fig.~\ref{fig:res_final_fmnist}) achieves an accuracy of 92.15\% in 51 epochs. This solution improved the latency (-79.38\%), energy (-88.43\%), and memory footprint (-63.05\%) compared to both the CapsNet and DeepCaps architectures, with almost the same accuracy as the last one, which is 93.94\%. 

\vspace*{-4pt}
\subsubsection{NASCaps for the SVHN Dataset}\ \\
The set of experiments for the SVHN dataset produced a solution (see pointer \rpoint{7} in Fig.~\ref{fig:res_final_svhn}) that reached an accuracy of 93.17\%, i.e., 3.52\% lower than the DeepCaps, in 56 epochs of training. This solution also significantly reduced the energy by 97.05\% and latency by 29.56\%, compared to the DeepCaps, but it requires 1.6x more memory. On the other hand, another interesting solution (see pointer \rpoint{8}) reached an accuracy of 92.53\%, while requiring 30.59\% lower energy, 59.63\% lower latency and 62.70\% lower memory, compared to the DeepCaps.

\vspace*{-4pt}
\subsubsection{NASCaps for the CIFAR-10 Dataset}\ \\
A solution found by the CIFAR10-NAS (see pointer \rpoint{9} in Fig.~\ref{fig:res_final_cifar}) achieved an accuracy of 85.99\% after 300 epochs of training, while significantly improving all the other objectives, compared to the DeepCaps architecture. This particular solution (\textit{NASCaps-C10-best} in Table~\ref{tab:cifar_complex}) reduced the energy consumption by 52.12\%, the latency by 64.34\% and the memory footprint by 30.19\%, compared to the DeepCaps executed on the CapsAcc accelerator, while encountering a slight accuracy drop of about 1\%, while using the same training settings. Table~\ref{tab:cifar_complex} reports also other Pareto-optimal DNN architectures found by our \textit{NASCaps} framework for the CIFAR-10 dataset. 

\begin{table}[ht]
\caption{Selected CIFAR-10 architectures after 300-epoch training.}% \red{I commented some more architectures to reduce size}}% \vm{I suggest to include some numbers - this table shows the CIFAR-10 after 300 epochs (approx 24 hours of training) instead of Fig~\ref{fig:cifar_complex} the architectures that are worser than CapsNet are filtered out; you can also remove some not-interesting points}}

\label{tab:cifar_complex}
\resizebox{.9\columnwidth}{!}{
\begin{tabular}{lrrrr}
\toprule
           \textbf{Architecture} &  \textbf{Accuracy} &   \textbf{Energy} & \textbf{Latency} &  \textbf{Memory} \\
\midrule
        DeepCaps~\cite{deepcaps} & 87.10\%\tablefootnote{\label{note_acc}Note: The accuracy reported for the DeepCaps and CapsNet do not 100\% match with the ones reported in~\cite{deepcaps}. This can be attributed to the differences in the training hyper-parameter setup, as their papers do not disclose the complete in-depth information about the training that can ensure reproducibility of their results.} & 36.30 mJ & 4.29 ms &  9,052 kiB \\
 NASCaps-C10-best \rpoint{9} & 85.99\% & 17.38 mJ & 1.53 ms &  6,319 kiB \\
 NASCaps-C10-a0d & 74.11\% &  4.53 mJ & 1.12 ms &  1,718 kiB \\
 NASCaps-C10-9fd & 74.00\% &  5.11 mJ & 0.36 ms &    713 kiB \\
 NASCaps-C10-658 & 73.91\% &  5.06 mJ & 1.54 ms &  5,573 kiB \\
 %NASCaps-C10-458 & 73.82\% & 11.57 mJ & 1.12 ms &  1,740 kiB \\
 %NASCaps-C10-51d & 73.03\% & 11.99 mJ & 0.80 ms &  1,748 kiB \\
 % NASCaps-C10-9c9 & 69.50\% &  1.41 mJ & 2.98 ms & 14,579 kiB \\
 %NASCaps-C10-6c2 & 67.15\% &  0.02 mJ & 0.10 ms &    422 kiB \\
         CapsNet~\cite{sabour} & 55.85\%\textsuperscript{\ref{note_acc}} & 88.80 mJ & 1.82 ms &  8,573 kiB \\
% NASCaps-C10-e03 & 49.32\% &  0.10 mJ & 0.69 ms &    360 kiB \\
% NASCaps-C10-f89 & 49.19\% &  0.03 mJ & 0.14 ms &    422 kiB \\
% NASCaps-C10-579 & 48.35\% &  0.09 mJ & 0.65 ms &    361 kiB \\
% NASCaps-C10-02c & 10.02\% & 15.11 mJ & 1.33 ms &  1,779 kiB \\
% NASCaps-C10-fe9 & 10.00\% &  0.08 mJ & 0.65 ms &    360 kiB \\
% NASCaps-C10-eb4 & 10.00\% &  4.97 mJ & 0.80 ms &  1,731 kiB \\
\bottomrule
\end{tabular}}
\vspace*{-4pt}
\end{table}

\subsubsection{Transferability of the Selected DNNs Across Different Datasets}
To test the transferability of the DNN solutions found by our \textit{NASCaps} framework, the dataset-specific found DNNs have been also trained and tested on the rest of the considered datasets. Table~\ref{tab:transferability} reports the matrix of highest-accuracy solutions, obtained by this transferability analysis. 

The \textit{NASCaps-C10-best} architecture of Table~\ref{tab:cifar_complex} resulted also particularly accurate for the other datasets. For the MNIST dataset, it achieved an accuracy of 99.72\% in 37 epochs of training, which is also higher than the solutions found by the MNIST-NAS. 
For the Fashion-MNIST dataset, it reached an accuracy of 93.87\% in 32 epochs of training, which is even higher than the DeepCaps after 100 epochs of training. When tested on the SVHN dataset, it reached an accuracy of 96.59\%, thus outperforming the highest-accuracy DNN found during the SVNH-NAS. The \textit{NASCaps-C10-best} architecture is similar to the DeepCaps, but it has two initial convolutional layers and three CapsCell blocks, without skip connection. 
The highest-accuracy architecture found by the MNIST-NAS also proved to work well with the Fashion-NMIST dataset, reaching an accuracy of 93.34\% after 91 epochs of training.

\begin{table}[h]
\caption{Highest-Accuracy DNNs found by the dataset-specific NAS, which are then trained for the other datasets for 100 epochs. %Accuracies of best NAS-found architectures on different datasets (100 epochs)
%\red{Some missing data.. 5 NNs to be trained. Should this be included?} \blue{If we have time yes.}\red{I wrote some temporary data with the epochs at which the training completed to this point.}
} 
\label{tab:transferability}
\vspace*{-2pt}
\resizebox{0.95\columnwidth}{!}{
\begin{tabular}{lcccc}
\toprule
\textbf{Architecture} &
  \multicolumn{1}{c}{\textbf{MNIST}} &
  \multicolumn{1}{c}{\textbf{FMNIST}} &
  \multicolumn{1}{c}{\textbf{SVHN}} &
  \multicolumn{1}{c}{\textbf{CIFAR-10}} \\ \midrule
\textbf{NASCaps-MNIST-best \rpoint{4}}    & 99.65\% & 93.34\% & 96.36\% & 71.44\% \\
\textbf{NASCaps-FMNIST-best \rpoint{6}}   & 99.49\% & 92.15\% & 93.12\% & 68.34\%   \\
\textbf{NASCaps-SVHN-best \rpoint{7}}     & 99.51\% & 91.43\%  & 93.17\% & 63.72 \%       \\
\textbf{NASCaps-C10-best \rpoint{9}} & \textbf{99.72\%} & \textbf{93.87\%}  & \textbf{96.59\%} & \textbf{76.46\%} \\ \bottomrule
\end{tabular}}
\vspace*{-4pt}
\end{table}

The results reported in Table \ref{tab:transferability} show how the solution \textit{NASCaps-C10} is the best overall architecture found during the four searches performed. This is due to multiple reasons: the evolutionary process was based on a random initial parent population that has been newly generated at each search. Moreover, the small size of the initial parent population may have contributed to a non-convergence of the four dataset-specific searches that have been performed. Also, not each one of the four searches reached the same generation at the end of the experiments.  

\vspace*{-8pt}
\subsection{Summary of Key Results}
%@andrea, revised by @alberto

The above results show how our \textit{NASCaps} framework has been able to explore multiple solutions with diverse tradeoffs, thanks to the usage of an evolutionary algorithm for a multi-objective search. It has been possible to generate and test 690 candidate networks for the four dataset-specific searches. 
Using four high-end NVIDIA Tesla V100-SXM2, our \textit{NASCaps} framework required 90 GPU-hours to test the partially-trained candidate networks. The new 64 Pareto-optimal architectures have been fully-trained, requiring in total additional 682 GPU-hours (i.e., 28 days). 
Our approach allowed to outperform many objectives of the SoA solutions when performing the full-training, despite the strict time constraints applied to the single searches. In summary, our framework %and reduced-training approach 
allowed to:
\vspace*{-3pt}
\begin{itemize}[leftmargin=*]
    \item Derive some interesting architectures, such as the above-discussed \textit{NASCaps-C10-best} that reached an almost similar accuracy as of the SoA, while significantly improving all other objectives of the search, i.e., energy, memory and latency.
    \item Perform early candidate selection while still achieving high accuracy after performing the full training.
    \item Achieve good transferability between different datasets, as demonstrated by the fact that the \textit{NASCaps-C10-best} DNN, which is found for the CIFAR10-specific search, outperforms other dataset-specific searches also on other datasets.
\end{itemize}
\vspace*{-3pt}

\vspace*{-8pt}
\section{Conclusion}
%@alberto

In this paper, we presented \textit{NASCaps}, a framework for the Neural Architecture Search (NAS) of Convolutional Capsule Networks (CapsNets). The set of optimization goals for our framework are the network accuracy and the hardware efficiency, expressed in terms of energy consumption, memory footprint, and latency, when executed on the specialized hardware accelerators. We performed a large-scale NAS using GPU-HPC nodes with multiple Tesla V100 GPUs, and found interesting DNN solutions that are hardware-efficient yet highly-accurate, when compared to SoA solutions. Our framework is even more beneficial when the design times are short, training resources at the design center are limited, and the DNN design is subjected to short training durations. Our \textit{NASCaps} framework can ease the deployment of DNNs based on capsule layers in resource-constrained IoT/Edge devices. We will open-source our framework at \mbox{\url{https://github.com/ehw-fit/nascaps}}.

\vspace*{-8pt}
\begin{acks}

This work has been partially supported by the Doctoral College Resilient Embedded Systems which is run jointly by TU Wien's Faculty of Informatics and FH-Technikum Wien, and partially by Czech Science Foundation project GJ20-02328Y.
% Computer cluster:
The computational resources were supported by The Ministry of Education, Youth and Sports from the Large Infrastructures for Research, Experimental Development and Innovations project ``e-Infrastructure CZ – LM2018140''.

\end{acks}

\vspace*{-2pt}
\bibliographystyle{abbrvnat}
\bibliography{main}

\end{document}